%% file: acl_latex.tex
\pdfoutput=1

\documentclass[11pt,dvipsnames,table]{article}

\usepackage{acl}
\usepackage{enumitem}
\usepackage{times}
\usepackage{latexsym}
\usepackage{algorithm}
\usepackage{algpseudocode}
\usepackage{mathtools}
\usepackage{amssymb}
\usepackage{fancyvrb}
\usepackage{fancyhdr}
\usepackage{makecell}
\usepackage{booktabs}
\usepackage{amsmath}
\usepackage{listings}
\usepackage{comment}
\usepackage{hyperref}

\newcommand*{\up}{\cellcolor{cyan!25}}
\newcommand*{\dn}{\cellcolor{orange!35}}

\algrenewcommand\algorithmicrequire{\textbf{Input:}}
\algrenewcommand\algorithmicensure{\textbf{Output:}}

\usepackage[T1]{fontenc}

\usepackage[utf8]{inputenc}

\usepackage{microtype}

\title{TV-TREES: Multimodal Entailment Trees \protect\\ for Neuro-Symbolic Video Reasoning}

\author{Kate Sanders \ \ \ \ \ Nathaniel Weir \ \ \ \ \ \  Benjamin Van Durme\\
  Johns Hopkins University \\
  \texttt{\{ksande25, nweir, vandurme\}@jhu.edu} }

\begin{document}
\maketitle

\input{content/00_abstract}
\input{content/01_introduction}
\input{content/02_related_work}
\input{content/03_task}
\input{content/04_method}
\input{content/05_eval}

\input{content/06_exps}
\input{content/07_conclusion}
\input{content/08_limitations}

\bibliography{anthology,custom}
\bibliographystyle{acl_natbib}

\clearpage
\appendix

\input{content/appendix}

\end{document}

%% file: content/00_abstract.tex
\begin{abstract}
\label{sec:abs}
It is challenging for models to understand complex, multimodal content such as television clips, and this is in part because video-language models often rely on single-modality reasoning and lack interpretability. To combat these issues we propose TV-TREES, the first multimodal entailment tree generator. TV-TREES serves as an approach to video understanding that promotes interpretable joint-modality reasoning by searching for trees of entailment relationships between simple text-video evidence and higher-level conclusions that prove question-answer pairs. We also introduce the task of multimodal entailment tree generation to evaluate reasoning quality. Our method's performance on the challenging TVQA benchmark demonstrates \emph{interpretable}, state-of-the-art zero-shot performance on full clips, illustrating that multimodal entailment tree generation can be a best-of-both-worlds alternative to black-box systems.
\end{abstract}

%% file: content/01_introduction.tex
\section{Introduction}
\label{sec:intro}
\begin{figure}[ht!]
  \includegraphics[width=.49\textwidth]{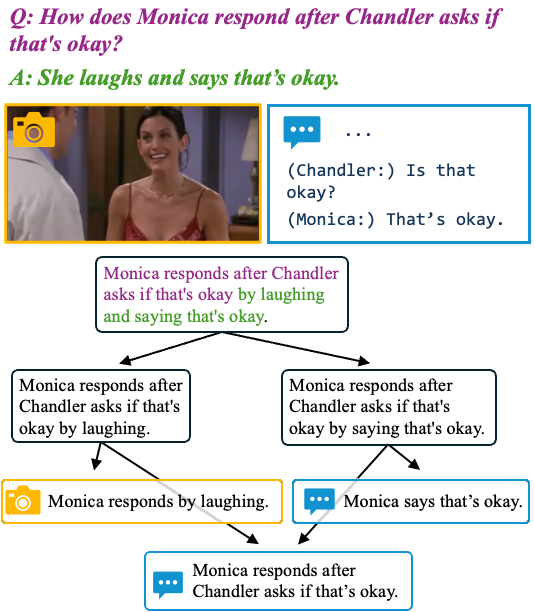}
  \caption{A QA pair and corresponding video clip and dialogue from the TVQA dataset~\cite{lei2018tvqa}, and a multimodal entailment tree, recursively produced by our approach (top-down). Trees are created by recursively retrieving atomic evidence from the transcript and video frames and decomposing the QA pair into compositionally equivalent hypotheses until each can be directly entailed by the retrieved evidence.}
\end{figure}

Videos account for a large portion of content available and consumed online, but automated reasoning over semantically complex video-language data remains a challenging and under-explored problem. A popular task for assessing models' video understanding is narrative-centric video-language question-answering: Given a natural language question, a video clip of a movie or TV show, and a corresponding dialogue transcript, the goal is to return a correct natural language answer to the question using the video-language data.

Methods tackling the video-language QA task~\cite{yang2022zero,li2020hero,ko2023large} frequently take the form of large, joint-modality transformer models. 
Analyses suggest their ability to perform joint visual-language reasoning is limited, as they often rely on either text or visual content but not both~\cite{rawal2023revealing}. Assessing modality reliance and reasoning quality overall is difficult given their lack of interpretability: 
While LLMs now facilitate increasingly transparent explanation generation alongside outputs~\cite{zhao2023explainability}, video-language models generally lack this ability.

Entailment trees~\cite{dalvi-etal-2021-explaining}, or trees of entailment relationships between atomic premises and higher-level conclusions, have been shown to serve well as the structural basis for text-only QA tasks by systematically and transparently modeling logical reasoning chains~\cite{weir2023dynamic}. We embrace this approach: We develop (1) the first multimodal entailment tree generator, TV-TREES (the \textbf{T}ransparent \textbf{V}ideo\textbf{-T}ext \textbf{RE}asoning with \textbf{E}ntailment \textbf{S}ystem), and (2) the \textit{task} of multimodal entailment tree generation to assess the reasoning ability of such systems.

In contrast to existing black-box QA systems, TV-TREES focuses on the search and manipulation of atomic ``facts'' retrieved from video clips to search for proofs for video-language question-answer pairs. The approach jointly reasons over both modalities and, crucially, the resulting entailment trees provide human-interpretable evidence and natural language explanations for each logical operation, enabling direct analysis of the model's underlying reasoning. Our entailment tree evaluation method builds on work in informal logic, adapting these ideas to the multimodal domain with an emphasis on reliable evaluation. 

We show that our multimodal reasoning system performs competitively on zero-shot video-language QA for the difficult TVQA dataset~\cite{lei2018tvqa} while \textit{simultaneously providing} comprehensive and interpretable reasoning traces. Further, TV-TREES achieves state-of-the-art performance using full-length clips as input.

In summary, our contributions are:
\vspace{-1mm}
\begin{enumerate}[leftmargin=*]
\setlength\itemsep{0em}
\item The first multimodal entailment tree generator, an explainable video-language understanding system that emphasizes logical reasoning across modalities by searching for proofs for question-answer pairs.
\item The task of multimodal entailment tree generation for evaluating step-by-step video-language reasoning quality.
\item Results demonstrating state-of-the-art zero-shot video-language QA performance on TVQA, ablation experiments demonstrating the benefit of joint-modality reasoning, and quantitative and qualitative analyses of entailment tree quality.
\end{enumerate}

%% file: content/02_related_work.tex
\section{Related Work}
\label{sec:relwork}
\subsection{VideoQA}
\label{sec:r-videoqa}
QA over images makes up a large portion of multimodal question-answering work~\cite{zou2020survey}. VideoQA benchmarks constitute a smaller portion of this area \cite{zhong2022video} and often focus on simple content and questions \cite{jang2017tgif}, but some recent VideoQA datasets have targeted models' commonsense knowledge and inference ability, namely TVQA and MovieQA \cite{lei2018tvqa, tapaswi2016movieqa}. We focus our experiments on TVQA as evidence suggests about half of MovieQA's questions can be answered with the question and answer options alone. \cite{jasani2019we}

Recently, vision-and-language transformers have substantially improved performance on these VideoQA tasks \cite{zhou2020unified}, and can often reason over complex content without an external knowledge base~\cite{kim2021vilt, wang2021distilled, salin2022vision}. In contrast to these video-language models, \citet{khurana2021video} highlight alternative strategies for VideoQA such as attention-free methods, attention-based methods, memory network methods, and hierarchical reinforced methods. Notably, \citet{zhao2018open, zhao2020open} propose a hierarchical encoder-decoder model that uses adaptive video segmentation based on the question contents.

\subsection{Explainable Multimodal Understanding}
Traditional techniques like kernel visualization and perturbation have been considered for video explainability \cite{hiley2019explainable,li2021towards} alongside other approaches that consider low-level reasoning steps for simple tasks  \cite{zhuo2019explainable, roy2019explainable, nourani2020don}. Additionally, \citet{lu2022learn} introduce a transparent reasoning benchmark for vision-text QA. The approaches most similar to our work are \cite{chen2021explainable}  and \cite{mao2022dynamic}. \citet{chen2021explainable} ground relevant textual entities in video and dialogue through a heatmap over the input as an explanation for the produced output. Our work differs in that we show exactly what data pieces contribute to the final output, explicitly model each step of the reasoning process, and don't require fine-tuning on the target dataset or domain. \citet{mao2022dynamic} uses a chain-of-thought explanation system based on a video scene graph to answer questions about actions and objects in short video clips and GIFs. This method does not consider dialogue and focuses on simple visual questions, instead of complex inferential reasoning that TV-TREES tackles. Furthermore, the input for their proposed system only spans a few seconds.

\subsection{Entailment Tree Generation}
This paper draws inspiration from recent work on constructing natural language entailment trees to explain reasoning. The notion starts with 
\citet{dalvi-etal-2021-explaining}, who introduce an expert-annotated dataset of compositional trees showing how a hypothesis follows as a logical consequence of a series of multi-premise entailment steps starting from verified support facts. 
More recent work has introduced methods to tackle Dalvi et al.'s reconstruction task~\cite{bostrom-etal-2022-natural, neves-ribeiro-etal-2022-entailment} and to use entailment trees as a basis for neuro-symbolic reasoning~\cite{tafjord-etal-2022-entailer, weir2023dynamic}. Our work is most similar to \citet{weir2023dynamic}, who introduce a QA system that reasons by searching for entailment trees grounded in a knowledge source. We extend this notion to the multimodal setting and address the resulting challenges.

\subsection{Multimodal Entailment}
There is a selection of work that considers entailment in images and video: \cite{xie2019visual} introduce a dataset of image-entailment pairs similar to the SNLI \cite{bowman2015large} corpus, and \cite{do2020snli} add natural language explanations to the pairs. More specific visual entailment tasks in this domain have been proposed as well \cite{thomas2022fine, li2023scene}., and \cite{suzuki2019multimodal} introduce a logic system for identifying entailment between images and captions. \citet{liu2020violin} introduce VIOLIN, a dataset of videos paired with natural language inferences that are entailed or contradicted by the video content, and traditional models \cite{li2020hero,sun2022long} as well as tailored approaches \cite{li2021adaptive, chen2021explainable} are trained for the task.

%% file: content/03_task.tex
\section{Multimodal Entailment Trees}
\label{sec:task}
\begin{figure*}[!ht]
  \includegraphics[width=\textwidth]{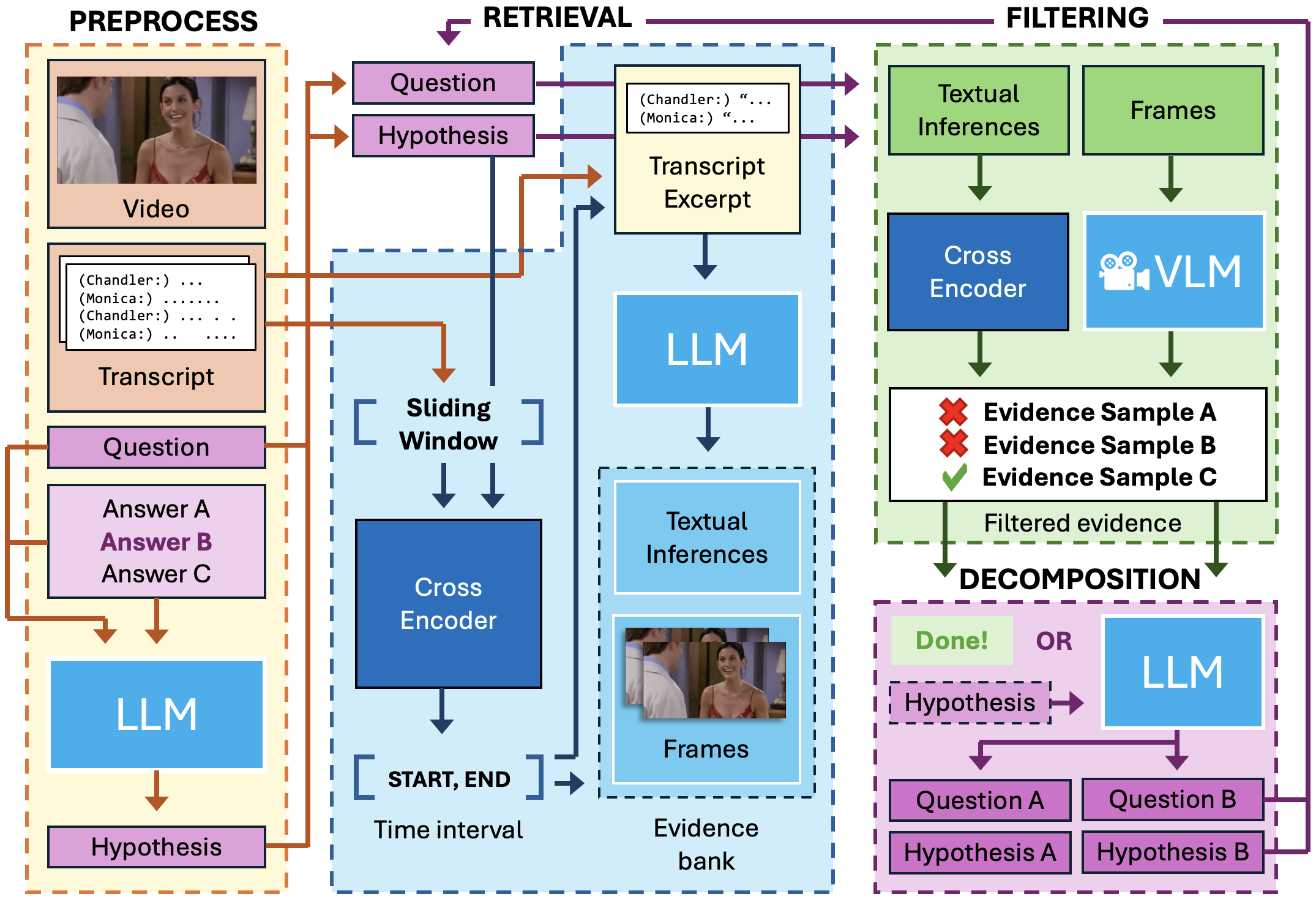}
  \caption{The TV-TREES pipeline. The system searches for evidence in the video clip and transcript that it can use to prove higher-level conclusions, with the goal of recursively constructing a tree of entailment relationships between these conclusions and low-level evidence. The figure highlights the system's three primary modules: Retrieval, filtering, and decomposition. \textbf{Retrieval} involves identifying the best time interval of the video to sample from and then extracting natural language inferences and video frames from the input data. \textbf{Filtering} involves filtering the extracted data samples with NLI classifiers and VQA systems to identify evidence that proves the answer to the question. \textbf{Decomposition} addresses when no evidence is found--- an LLM is used to decompose the question and hypothesis into two smaller sub-claims, to be each recursively proven through the same process.}
  \label{fig:pipeline}
\end{figure*}

We now introduce the task of multimodal entailment tree generation for video-language QA.

\subsection{Task formulation}

\textbf{Input} \hspace{2mm} Following \citet{dalvi-etal-2021-explaining}, as input we consider hypothesis $h_{(q,a)}$ (the declarative form of a question-answer pair) and an evidence bank. Traditionally, the evidence bank takes the form of a set of natural language sentences, but in the multimodal domain it is a video clip $V$ and corresponding dialogue transcript $D$, treated as sets of video frames and dialogue lines, respectively. 

\paragraph{Output}
We define entailment trees as recursive structures which take the form $T:=(h,e)$. $h$ is a hypothesis and $e$ is evidence. $e$'s form is either a
\begin{enumerate}
\setlength\itemsep{0em}
    \item \textit{Leaf}: A (possibly empty) subset of items from evidence bank $\{V\cup D\}$.
    \item \textit{Branch}: A pair of two distinct entailment subtrees $T_1:=(h_1,e_1)$ and $T_2:=(h_2,e_2)$.
\end{enumerate}

\noindent Leaves with empty evidence sets are labeled \textit{null}.

The purpose of an entailment tree is to illustrate the compositional reasoning necessary to reach a conclusion from an initial evidence bank using entailment relationships. In a \textit{well-formed tree}, the evidence $e$ in any tree node $(h,e)$ must explicitly entail the hypothesis $h$. For a leaf node, $e$ entails $h$ if a human would reasonably infer that $h$ is true given evidence $e\subseteq \{V\cup D\}$. For a branching node, $e$ entails $h$ if a human would reasonably infer that $h$ is true given hypotheses $h_1$ and $h_2$.

\paragraph{Objective}
Given inputs $h_{(q,a)},V,\text{ and }D$, our objective is to return a well-formed entailment tree $T$ that includes null leaves if and only if $a$ is not a correct answer to question $q$.

\subsection{Evaluation}
\label{sec:t-eval}
To serve as a second, distinct objective from raw QA performance, we propose an evaluation method for assessing the reasoning quality of multimodal entailment trees inspired by work in compositional entailment evaluation \cite{weir2024enhancing}. Informal logic theory posits that natural language arguments may be evaluated in terms of their \textit{acceptability}, \textit{relevance}, and \textit{sufficiency}~\cite{johnson-blair-1977-logical}, and we consider each node in an entailment tree as an ``argument" to be scored using these qualia. Below, we formulate them through an information-theoretic lens to establish a set of evaluation metrics. We use the Shannon definition of information gain,
$I(x\hspace{1mm}|\hspace{1mm}y)=-\log P(x\hspace{1mm}|\hspace{1mm}y),$
where $P(x)$ is the probability that natural language statement $x$ is true conditioned on natural language statement(s) $y$.

\paragraph{Acceptability} Hypotheses at every node should be complete and verifiable natural language statements that are understandable to a human, and hypotheses at leaf nodes should be factually accurate statements conditioned on the world state $\{V\cup D\}$. These items may be formalized as
\begin{gather}I(h)\in [0,1]\hspace{2mm}\forall h\in T\\
I(h\hspace{1mm}|\hspace{1mm}V\cup D)=0\hspace{2mm}\forall {h\in T_{\text{leaves}}}.\end{gather}

\noindent\textbf{Relevance} \hspace{2mm} For each branching node $T_0:=(h_0,(T_1,T_2))$, hypotheses $h_1$ and $h_2$ should both be \textit{conditionally relevant} to $h_0$, meaning that they each introduce distinct information that contributes to the compositional entailment of $h_0$. Formally, this metric is met if
\begin{gather}I(h\hspace{1mm}|\hspace{1mm}h_1,h_2)<I(h\hspace{1mm}|\hspace{1mm}h_2)\hspace{2mm}\forall (h,e)\in T_{\text{branches}}\\
I(h\hspace{1mm}|\hspace{1mm}h_1,h_2)<I(h\hspace{1mm}|\hspace{1mm}h_1)\hspace{2mm}\forall (h,e)\in T_{\text{branches}}\end{gather}

\noindent\textbf{Sufficiency} \hspace{2mm} For each branching node $T_0:=(h_0,(T_1,T_2))$, hypotheses $h_1$ and $h_2$ should compositionally entail $h_0$, or
\begin{gather}I(h_0\hspace{1mm}|\hspace{1mm}h_1,h_2)=0\hspace{2mm}\forall (h_0,(T_1,T_2))\in T.\end{gather}

\noindent Given these metric formulations, we explore practical implementations of them in Section~\ref{sec:eval}.

%% file: content/04_method.tex
\section{TV-TREES}
\label{sec:method}
We now introduce our multimodal entailment tree generator, pictured in \autoref{fig:pipeline}.

\subsection{System overview}
\label{sec:m-overview}
TV-TREES is a recursive search algorithm that involves three primary procedures:
\vspace{-1mm}
\begin{enumerate}[leftmargin=*]
\setlength\itemsep{0em}
\item \textbf{Retrieval\hspace{2mm}} Given a hypothesis and evidence bank, the system samples evidence candidates from the bank that may sufficiently entail the current hypothesis.
\item \textbf{Filtering\hspace{2mm}} The system tests whether any retrieved evidence entails the hypothesis. If such evidence was retrieved, it is returned and the current tree node becomes a leaf.
\item \textbf{Decomposition\hspace{2mm}} If the previous steps result in insufficient evidence, the system decomposes the hypothesis into two sub-hypotheses such that proving both independently is equivalent to proving the original hypothesis. The process is recursively called using these sub-hypotheses.
\end{enumerate}

The interaction of these three parts is illustrated in Algorithm 1. Given a hypothesis $h$, transcript sample $D'\subseteq D$, and video sample $V'\subseteq V$, the system first returns evidence from the transcript relevant to $h$ (line 1) and identifies whether any of it entails $h$ (2). If such evidence was retrieved, $e$ is set to the best evidence (3-4), and the leaf node is returned (17). Otherwise, $h$ is decomposed into sub-hypotheses $h_0$ and $h_1$ (8) and the algorithm is recursively called on these newly constructed sub-problems (9-10), treating the generated sub-proofs as explanation $e$ (11). If the maximum depth is reached during recursion, the evidence at that node is set to the empty set (5-6). If textual evidence cannot be found for the current node nor any of the downstream nodes (14), then the visual evidence in sample $V'$ is sampled, filtered, (13) and assigned to $e$ where applicable (15) in the same manner as the text content.
Below, we detail the implementation of the subroutines called by Algorithm 1.

\begin{algorithm}[t!]
\label{alg:generate}
\caption{Tree generation, \textproc{Generate}}
\begin{algorithmic}[1]
\Require Hypothesis $h$, transcript sample $D'\subseteq D$, video sample $V'\subseteq V$, current depth $k$
\Ensure Tree candidate $\hat{T}\coloneqq (h,p')$
\State $F_D \gets \textproc{Retrieve}(D'\hspace{1mm} | \hspace{1mm}h)$
\State $F_D' \gets \textproc{Filter}_{\hspace{.5mm}D}(F,h)$
\If{$F_D' \neq \emptyset$}
\State $e \gets \textproc{Best}_{\hspace{.5mm}D}(F_D'\hspace{1mm} | \hspace{1mm}h)$
\ElsIf{$k \geq k'$}
\State $e \gets \emptyset$
\Else
\State $h_0, h_1 \gets \textproc{Decompose}(h\hspace{1mm}|\hspace{1mm} T')$
\State $T_0 \gets \textproc{Prove}(h_0, D', V', k+1)$
\State $T_1 \gets \textproc{Prove}(h_1, D', V', k+1)$
\State $e \gets (T_0,T_1)$
\EndIf
\State $F_V' \gets \textproc{Filter}_{\hspace{.5mm}V}(V'\hspace{1mm} | \hspace{1mm}h)$
\If{$\textproc{Null}(e)$ and $F_V'\neq\emptyset$}
\State $e \gets \textproc{Best}_{\hspace{.5mm}V}(F_V'\hspace{1mm} | \hspace{1mm}h)$
\EndIf
\State \textbf{return} $(h,e)$
\end{algorithmic}
\end{algorithm}

\subsection{Preprocessing}
\label{sec:m-preprocessing}
\textbf{Hypothesis Generation} \hspace{2mm}
We first prompt GPT-3.5~\cite{ouyang2022training} to generate a single declarative statement that contains the full semantic meaning of an initial QA pair.\footnote{All LLM and VLM prompts are included in Appendix A.}

\paragraph{Evidence Localization}
Given the hypothesis, TV-TREES attempts to identify a relevant window of the video clip and transcript to sample evidence from. We use a cross-encoder model trained on the MS MARCO passage ranking task~\cite{bajaj2016ms}\footnote{\href{https://huggingface.co/cross-encoder/ms-marco-MiniLM-L-12-v2}{huggingface.co/cross-encoder/ms-marco-MiniLM-L-12-v2}} to rank six-line transcript passages on their computed similarity with the generated hypothesis. We use a sliding window to calculate scores for every potential sample and return the highest-scoring excerpt. If a sufficient window is identified, the vision pipeline inherits this same window. If no dialogue sample is found, the system uses all video frames as evidence and omits text entirely.

\begin{figure}[t!]
  \includegraphics[width=.49\textwidth]{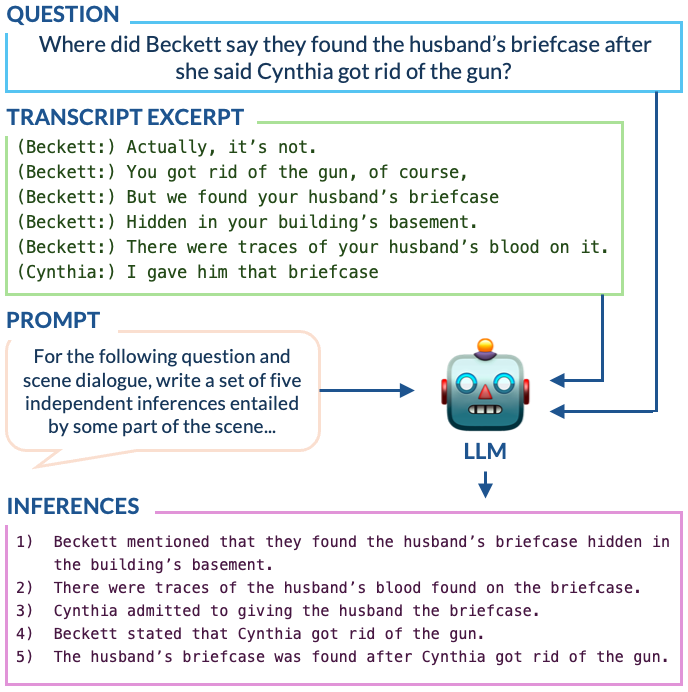}
  \caption{An example question from TVQA, corresponding dialogue excerpt sampled by TV-TREES, and set of inferences generated from these inputs.}
  \label{fig:inf}
\end{figure}

\subsection{Evidence Retrieval (\textproc{Retrieve})}
\label{sec:m-retrieval}
Existing natural language inference (NLI) models are not well-suited for classifying entailments within highly contextual and social dialogue, which often requires sophisticated inferential ability. Therefore, we use GPT-3.5 to generate a set of five natural language inferences about the dialogue, conditioned on a question form of the hypothesis, $q$, written in the style of a dataset like SNLI~\cite{DBLP:journals/corr/BowmanAPM15}. Presenting the question under discussion in the interrogative form significantly reduces the hallucination rate compared to passing in the original hypothesis. $q$ is also generated via GPT-3.5 taking the hypothesis $h$ as input. Example generated inferences are shown in \autoref{fig:inf}.

\subsection{Evidence Filtering (\textproc{Filter})}
\label{sec:m-filtering}
We use a cross-encoder trained on SNLI and MNLI~\cite{williams2017broad}\footnote{\href{https://huggingface.co/cross-encoder/nli-distilroberta-base}{huggingface.co/cross-encoder/nli-distilroberta-base}} to determine whether any of the retrieved inferences entail the hypothesis. We accept any sample that achieves a logit score above a given threshold.

We apply a secondary entailment filter using GPT-3.5 that ensures the inferences are accurate descriptions of the content presented in the dialogue. This is important because, while conditioning the inference generator on an interrogative form of the hypothesis mitigates hallucinations, it does not eliminate them entirely.

Finally, as the cross-encoder tends to ignore negation, we additionally pass the filtered inference-hypothesis pairs to a GPT-3.5 prompt to verify the entailment a final time. The system retains the inferences that pass all three filters.

\subsection{Decomposition (\textproc{Decompose})}
\label{sec:m-decomp}
If no retrieved evidence entails the current hypothesis, TV-TREES breaks down the hypothesis into two sub-hypotheses that are (1) complete sentences without ambiguous pronouns or decontextualized references and (2) compositionally equivalent to the original hypothesis, i.e., proving the two sub-hypotheses as true is approximately logically equivalent to proving the original hypothesis. We prompt GPT-3.5 to break the current hypothesis into two compositionally equivalent pieces of information. 

\subsection{Visual Reasoning (\textproc{Filter})}
\label{sec:m-vision}
We pass in the questions generated in Section~\ref{sec:m-decomp} alongside video frames from the localized evidence window (if applicable) into a vision-language model. In our experiments, we use LLaVA-7B~\cite{liu2023visual}.\footnote{The LLaVA-7B prompt is included in Appendix A.} To encourage conservative classifications, in addition to asking for ``yes" and ``no" answers we encourage the model to respond with ``not enough information" if it is unsure. If more than 10\% of the frames in the window result in an affirmative answer from the VLM model, the visual content is considered to contain sufficient entailing evidence, and the frame with the highest logits score is returned.

%% file: content/05_eval.tex
\section{Evaluation Methodology}
\label{sec:eval}
Traditionally, qualitative natural text evaluations have been conducted using humans~\cite{celikyilmaz2021evaluation}. 
Recently, researchers have considered whether these human evaluations could be replaced by high-performing LLMs like GPT-4~\cite{naismith2023automated}. We detail how we implement the evaluation metrics described in Section~\ref{sec:t-eval} with both human annotators and GPT-4. We report evaluation results for both methods in Section~\ref{sec:exps}.

\subsection{Human Evaluations}
\label{sec:e-human}
We evaluate trees using the metrics introduced in Section~\ref{sec:t-eval} (acceptability, relevance, and sufficiency) through three annotation tasks. The first task provides annotators with a tree's leaf node evidence (images or text) and asks them to assess the correctness of the leaf node hypotheses on a scale of 1-5 (\textbf{acceptability}) based on that evidence. The second task provides annotators with parent-child hypothesis $(h_0,h')$ pairs from branching nodes and asks if the child hypothesis $h'$ is relevant to the parent $h_0$ (\textbf{relevance}). The third task provides annotators with a full hypothesis triplet $(h_0,h_1,h_2)$ from a branching node with parent $h_0$ and child hypotheses $h_1$ and $h_2$ and asks (1) whether $h_1$ and $h_2$ each introduce distinct information (the other facet of \textbf{relevance}, we also call this \textbf{distinctness} for disambiguation purposes), and (2) if $h_0$ introduces information not provided by $h_1$ and $h_2$ together, to check for entailment (\textbf{sufficiency}). Through these tasks, annotators are also asked to indicate if any of the hypotheses or premises are malformed or otherwise uninterpretable (also \textbf{acceptability}). 

Every node in a multimodal entailment tree is assigned a binary score for each assessment described above (except for the correctness checks, which are collected on a scale of 1-5). We include all task instructions and layouts in Appendix D.

\subsection{GPT Evaluations}
\label{sec:e-gpt}
We take the qualia outlined in Section~\ref{sec:t-eval} and write three GPT-4 prompts testing (1) acceptability of evidence in the text domain, (2) acceptability of evidence in the vision domain, and (3) relevancy and sufficiency.\footnote{These prompts are included in full in Appendix E.} We use the same scoring values used in the human evaluations.

\subsection{Tree Scoring Paradigm}

We consider the mean normalized score of the three main evaluation qualia across all nodes as the overall ``composition score" for each individual tree, $$S=\frac{1}{3}(a+s+0.5(d+r))$$ where $a$ is the tree's mean normalized acceptability score, $d$ is the mean distinctness score, $r$ is the mean relevance score, and $s$ is the mean sufficiency score.

%% file: content/06_exps.tex
\section{Experiments}
\label{sec:exps}

We evaluate TV-TREES using the TVQA dataset as input data. We compare its zero-shot QA performance against competing video-language QA approaches to illustrate its practical usage, evaluate its overall tree quality through our evaluation method described in Section~\ref{sec:eval}, and organize its reasoning error modes through a qualitative study.

\paragraph{Setup}
We instantiate TV-TREES as it is described in Section~\ref{sec:method}, allowing for trees with up to 3 levels ($k=2$). Our experiments focus on the multiple-choice QA domain, and so we consider a question's correct answer to be the answer that results in a complete tree. In the case that the system does not successfully complete any tree for the five answer candidates, we consider the answer candidate with the "most complete" tree to be the correct answer, breaking ties by average entailment score. When complete trees are generated for multiple answers, we break ties in the same way.

\subsection{QA Evaluation}

We focus on video-language QA to take the first step adapting a text-only method to other domains, but complex video-language benchmarks are sparse: TVQA and MovieQA~\cite{tapaswi2016movieqa} are the two commonly used video-language datasets, but past research suggests that about half of MovieQA questions can be answered without reasoning over the video content \cite{jasani2019we}. Therefore, we focus our study on TVQA. 

\paragraph{Data} 
We evaluate our system on 3,000 multiple choice questions from the validation set of TVQA \cite{lei2018tvqa}. TVQA is a video-language QA benchmark that includes multiple choice questions about the dialogue and visual content of video clips taken from TV shows. The clips are about 60-90 seconds long and contain around 30 lines of dialogue. A sample question is shown in Figure 1.

\paragraph{Models}
In the zero-shot setting, we consider zero-shot systems FrozenBiLM~\cite{yang2022zero}, SeVILA \cite{yu2023self}, and VideoChat2~\cite{li2023mvbench}. We also include performance reported by other systems (not zero-shot) for context: STAGE~\cite{lei2019tvqa+}, HERO~\cite{li2020hero}, FrozenBiLM (fine-tuned)~\cite{yang2022zero}, and LLaMA-VQA~\cite{ko2023large}.

\begin{table}
    \centering
    \footnotesize
    \begin{tabular}{lccc}
    \toprule
        \textbf{Method} & \textbf{Transparent} & \textbf{Full Clips} & \textbf{TVQA}\\
        \midrule
        \multicolumn{4}{l}{\textbf{Fine-Tuned Methods}}\\
        \midrule
        STAGE  & \dn No & \up \textbf{Yes} & 70.5 \\
        HERO  & \dn No & \dn No & 74.2 \\
        FrozenBiLM  & \dn No & \dn No & 82.0 \\
        LLaMA-VQA  & \dn No & \dn No & \textbf{82.2} \\
        \midrule
        \multicolumn{4}{l}{\textbf{Zero-Shot Methods}}\\
        \midrule
        FrozenBiLM$^*$ & \dn No & \up \textbf{Yes} & 26.3 \\
        SeVILA  & \dn No & \up \textbf{Yes} & 38.2 \\
        VideoChat2 & \dn No & \up \textbf{Yes} & 40.6 \\
        TV-TREES$^\text{‡}$ & \up \textbf{Yes} & \up \textbf{Yes} & 44.9 \\
        \textbf{TV-TREES} & \up \textbf{Yes} & \up \textbf{Yes} & \textbf{49.4} \\
    \bottomrule
    \end{tabular}
    \caption{Table comparing vision-text understanding models on qualitative criteria and the TVQA benchmark. Experiment results suggest that TV-TREES and TV-TREES with text input only (TV-TREES$^\text{‡}$) outperform existing zero-shot methods on full clips. Competing method results are taken from their respective papers except for FrozenBiLM*, which we re-run on our validation subset with full clips as input. (On ground truth clip fragments, FrozenBiLM reports 59.7\% accuracy).}
    \label{tab:exps}
\end{table}

\paragraph{Ablations} Existing work notes that multimodal models are often biased toward the text modality, relying on text data for reasoning even for video-centric questions. To assess TV-TREES, we report TVQA performance conditioned on input modality. We compare system output when it is only provided with dialogue transcripts and then when it is only provided with video frames.

\paragraph{Results}
We report overall accuracy alongside qualitative comparisons between the approaches in \autoref{tab:exps}. As shown in the table, \textbf{TV-TREES outperforms existing zero-shot methods when using full clips}. The influence of the individual modalities on TV-TREES is further illustrated through the ablation experiment results in \autoref{fig:abl}, which reports the \% of questions which are correctly answered with complete trees and the \% that are correctly answered with incomplete proofs given text, visual, and multimodal evidence. The results show that \textbf{joint modality evidence improves both accuracy and correct tree completion in TV-TREES}.

\subsection{Tree Quality Evaluation}
\noindent\textbf{Setup}\hspace{2mm} We randomly sample 600 completed entailment trees generated by TV-TREES from the TVQA validation split, split evenly between evidence modality (text vs. multimodal) and tree complexity (ranging from one to seven tree nodes). We evaluate these sampled trees using the automatic GPT-4 approach as described in Section~\ref{sec:e-gpt}. We then sample 200 proofs from this set (evenly distributed across modalities and complexity) and evaluate this set with human annotators from Amazon Mechanical Turk as described in Section~\ref{sec:e-human}. For human annotations, we identify careful annotators through a preliminary pilot task where each annotator's work is scored by hand, and only high-scoring annotators are invited to annotate the full trees. More information regarding crowdsourcing is included in Appendix C.

\begin{figure}[t!]
    \centering
    \includegraphics[width=\columnwidth]{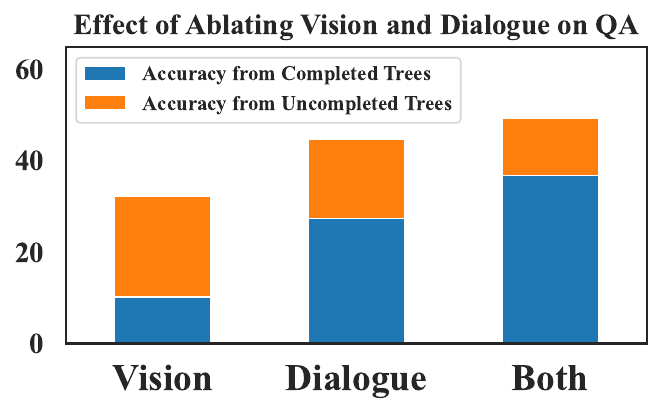}
    \caption{Ablation experiment results comparing TV-TREES performance on TVQA using only dialogue evidence, only visual evidence, and both modalities. We report the \% of questions answered correctly with completed trees and the \% answered correctly overall.}
    \label{fig:abl}
\end{figure}

\begin{figure}[t!]
    \centering
    \includegraphics[width=\columnwidth]{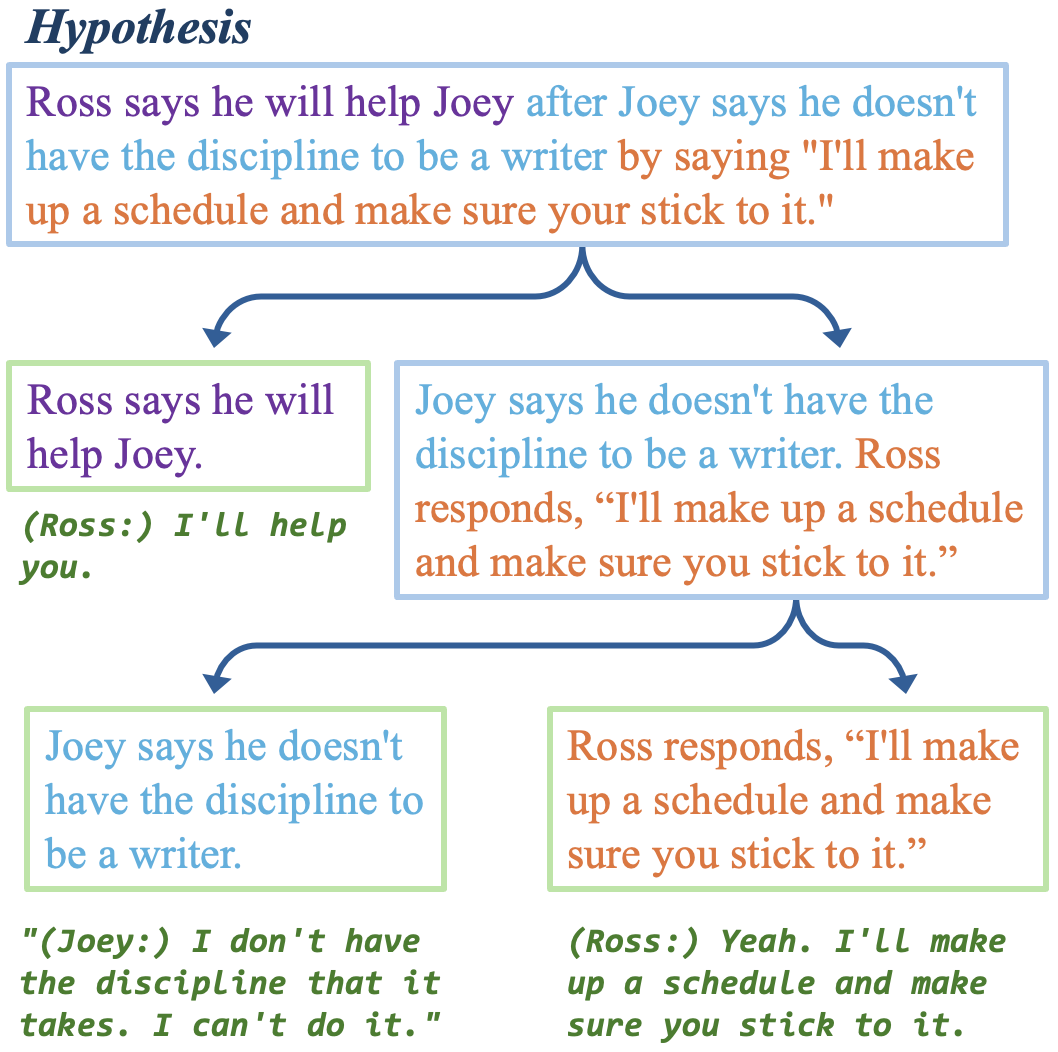}
    \caption{An example of a correct completed entailment tree produced with TV-TREES using text evidence (shown below the leaf nodes). Using our tree quality evaluation system detailed in Section 6.2, the tree earns perfect scores for acceptability (the dialogue entails the sub-hypotheses), relevance and distinctness (all child hypotheses help prove the parent and are distinct), and sufficiency (there is no information lost).}
    \label{fig:sample}
\end{figure}

\paragraph{Results}
We report results in Table~\ref{tab:eval}. For comparison, we include a high-quality tree produced by TV-TREES in Figure~\ref{fig:sample}. Generally, there is a close alignment between the machine scores and human scores, but GPT-4 tends to score the text-only trees more harshly than humans, and the multimodal trees more leniently. This is shown primarily in the resulting acceptability scores, and more moderately in the sufficiency scores. GPT-4 rated relevance scores more leniently for both modalities, which may stem from differences in human interpretations of the task instructions. Distinctness scores are almost identical between the two methods.

We find that the majority of error stems from acceptability issues. According to human evaluations, the vision module produces lower-quality inferences than the textual modules do. This is not surprising, as we are able to include additional entailment filters for the textual reasoning steps to remove lower-quality predictions before constructing the final entailment trees, whereas we do not have similar methods in place for visual inference. 

\subsection{Qualitative Analysis}
\paragraph{Setup}
Finally, we sample a set of 120 complete but incorrect entailment trees produced by TV-TREES on the TVQA dataset and analyze them to diagnose common error patterns. We find 8 main error classes, described below and in Table~\ref{tab:qual}.

\begin{table}
    \centering
    \footnotesize
    \setlength{\tabcolsep}{3pt}
    \begin{tabular}{lccccc}
    \toprule
        \textbf{Trees} & \textbf{Accept} & \textbf{Relev} & \textbf{Distinct} & \textbf{Suffic} & \textbf{Score} \\
        \midrule
        \multicolumn{6}{c}{\textbf{GPT-4 Evaluations}}\\
        \midrule
        \textbf{Text Only} & 58.4 & 99.6 & 87.7 & 88.6 & \textbf{74.3}       \\
        \textbf{Multimodal} & 61.0 & 99.6 & 90.6 & 93.9 & \textbf{77.8}       \\
        \textbf{All} & 59.7 & 99.6 & 89.1 & 91.2 & \textbf{76.0}       \\
        \midrule
        \multicolumn{6}{c}{\textbf{Human Evaluations}}\\
        \midrule
        \textbf{Text Only} & 65.6 & 93.9 & 88.8 & 93.6 & \textbf{78.9}      \\
        \textbf{Multimodal} & 51.8 & 98.1 & 91.2 & 92.8 & \textbf{72.9}       \\
        \textbf{All} & 58.7 & 96.0 & 91.7 & 93.2 &  \textbf{75.9}      \\
    \bottomrule
    \end{tabular}
    \caption{Entailment tree quality evaluations using human and LLM critics, reporting mean qualia scores alongside total score. 
    We partition results by modality: Trees using text evidence only, trees that use both modalities, and both groups combined (\textbf{all}). As shown, tree scores largely suffer due to acceptability, 
    highlighting
    the difficulty of extracting high-level inferences from dialogue and ambiguous video.}
    \label{tab:eval}
\end{table}

\paragraph{Failure Modes}
Visual reasoning errors are common among erroneous proofs, especially ones involving colors and character identification (we do not implement a character identification module in TV-TREES, so this error class is not surprising). Hallucinated text inferences is another common error class, occurring in the ``evidence retrieval'' module (Section~\ref{sec:m-retrieval}). Another common issue is the system ignoring negation in NL text: As documented in existing work \cite{hosseini2021understanding}, language models and classifiers often have difficulty recognizing negation in sentences. This can lead to specific entailment misclassifications. Finally, we notice that in some cases, the dataset question wording is difficult to interpret either due to coreference ambiguity or grammatical issues.

\paragraph{Results}
The distribution of error types among the sample set, reported in Table~\ref{tab:qual}, reflects and helps to explain the tree quality evaluation results reported in Table~\ref{tab:eval}: The most errors occur due to the acceptability of the produced evidence, as visual reasoning errors, hallucinated text inferences, and character and color identification account for 52\% of the tree errors in the qualitative study. Entailment misclassification and ignoring textual negation accounts for 26\% of the errors, explaining the lower sufficiency scores in Table~\ref{tab:eval}.

\begin{table}[t!]
    \centering
    \small
    \begin{tabular}{lcc}
    \toprule
        \textbf{Error Type} & \textbf{\%} & \textbf{Modality} \\
        \midrule
        Visual reasoning errors & 20\% & V \\
        Hallucinated text inferences & 19\% & T+V \\
        Entailment misclassification & 18\% & T+V \\
        Ignoring negation in text & 8\% & T+V \\
        Character identification & 7\% & V \\
        Ambiguous QA pairs & 7\% & T+V \\
        Color identification & 6\% & V \\
        Other & 15\% & \_ \\
    \bottomrule
    \end{tabular}
    \caption{Distribution of error modes across a sample of 120 complete but incorrect entailment trees generated by TV-TREES, analyzed by hand. Some of the error modes are particularly prevalent subclasses of other error modes, for example, ``color identification" could fall under ``general visual reasoning errors". (V = Occurs in vision/multimodal proofs only, T = occurs in both multimodal \textit{and} text-only proofs).}
    \label{tab:qual}
\end{table}

\begin{figure}[t!]
    \centering
    \includegraphics[width=\columnwidth]{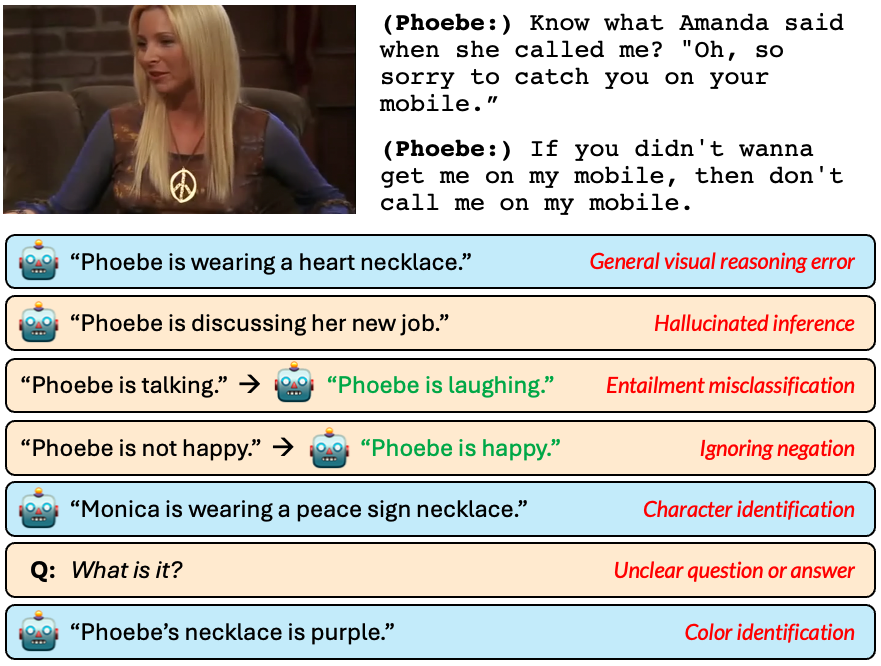}
    \caption{Examples of possible inference generation, entailment classification, and VQA filtering errors that illustrate the different failure mode categories identified in the qualitative analysis detailed in Section 6.3 and Table 3. Blue cells indicate vision-specific error types, and orange cells may occur in both text-only and multimodal proofs. Notably, "unclear question or answer" is not a failure of the system itself, but an artifact of the dataset used.}
    \label{fig:abl}
\end{figure}

%% file: content/07_conclusion.tex
\section{Conclusion}
\label{sec:conclusion}
We introduce the first neuro-symbolic entailment tree generator for multimodal content to improve the robustness, reliability, and interpretability of video-language understanding systems. We propose the \textit{task} of multimodal entailment tree generation for the assessment of generated tree reasoning quality, establishing an information-theoretic evaluation method grounded in informal logic theory. We show that our approach achieves state-of-the-art results on the zero-shot TVQA benchmark with full video clips, illustrating the potential for generated reasoning traces to improve downstream video-language understanding task performance. We show that interpretable, neuro-symbolic approaches to video understanding like TV-TREES are a strong alternative to existing methods, provide substantial new benefits, and highlight exciting directions for future research.

This paper is presented as an initial exploration into multimodal neuro-symbolic systems, and so there are many exciting avenues for future development and research. Individual components of this system could be improved for better performance on TVQA and related tasks - for instance, the vision querying system is fairly end-to-end, and semantically deconstructing the frames using a visual semantic role labeling model and using its outputs as evidence could result in a more sophisticated and transparent logical system. In future work, we also hope to explore the possibility of producing a collection of entailment trees pertaining to the same video clip and aggregating them to produce a comprehensive knowledge graph of the full video. We also hope to explore ways to improve computational efficiency and cost of using the system. Finally, there is significant room for future work in decomposing natural language text for entailment tree generation, and for establishing entailment between premises and grounded multimodal evidence.

%% file: content/08_limitations.tex
\section{Limitations}
\label{sec:lim}
We introduce an initial exploration into the task of multimodal entailment tree generation for video understanding, and so, there are inherent limitations that we hope to correct in future work. Most notably, our vision module underperforms compared to some systems - in future work, we hope to improve upon the existing end-to-end architecture as well as explore more compositional approaches. Furthermore, while we consider six lines of dialogue at a time to ensure sufficient context for textual inference, we do not do the same for visual analysis (instead working with only one frame at a time). Extending the immediate context for visual inference would likely improve performance as well. Finally, it is important to consider the domain that our system is used in, as model performance may vary in domains with limited dialogue, etc. We hope that this work inspires future research in this domain to improve upon our proposed pipeline. 

\section*{Acknowledgements}
This work has been supported in part by the U.S. National Science Foundation under grant NSF 2204926.

%% file: content/appendix.tex
\section{TV-TREES LLM Prompts}

We provide the LLM and VLM prompts used in the TV-TREES pipeline in Figures 9-16.\\




\section{Visual Prompt Anonymization Experiments}

We consider an additional component to the TV-TREES system outlined in Section~\ref{sec:method} that anonymizes any references to characters passed in to the visual entailment module. We pass any questions that will be used for visual QA prompts through a GPT filter that replaces any character names with common nouns and pronouns like ``the man", ``they", and ``the doctor". We report results below, comparing this alternate system to the competing methods and the standard TV-TREES method. We find that the anonymization paradigm results in a TVQA accuracy score of 48.1\% compared to the standard system's 49.4\%. We provide the anonymization GPT prompt in Figure 13 and a results table for comparison (Table 4).\\


\section{Amazon Mechanical Turk Details}

We evaluate generated tree quality through crowdsourced workers on Amazon Mechanical Turk with three main annotation tasks. We identify a separate group of quality annotators for each task by (1) setting the qualifications for the task to workers located within the United States with a HIT acceptance rate of 98\% and over 1000 completed HITS, and (2) running a pilot task with carefully selected questions to identify annotators who answer the preselected questions with high accuracy.

We estimate time completion for each version of the task uploaded to Mechanical Turk and set the payment values to an estimated \$15 per hour. No identifiable information of any annotators is present in this paper or in any artifacts we will release.\\

\section{Human Tree Evaluation Tasks}

Below, we include screenshots depicting the instructions and format of each task provided to annotators. We also include a table detailing the descriptions provided to annotators for each of the five acceptability scores (Table 5).\\

\section{GPT-4 Evaluation Prompts}
Prompts for GPT-4 evaluations are shown in Figures 17 - 19. Figure 17 shows the primary decomposition evaluation prompt, which accounts for relevancy, distinctness, and sufficiency. Figure 18 shows the textual acceptability for dialogue prompt, and Figure 19 shows the visual acceptability for screenshots prompt, which was passed to GPT-4V.\\

\onecolumn
\noindent\textbf{Acceptability:} See Figures 4 and 5.\\
\begin{figure}
  \includegraphics[width=\textwidth]{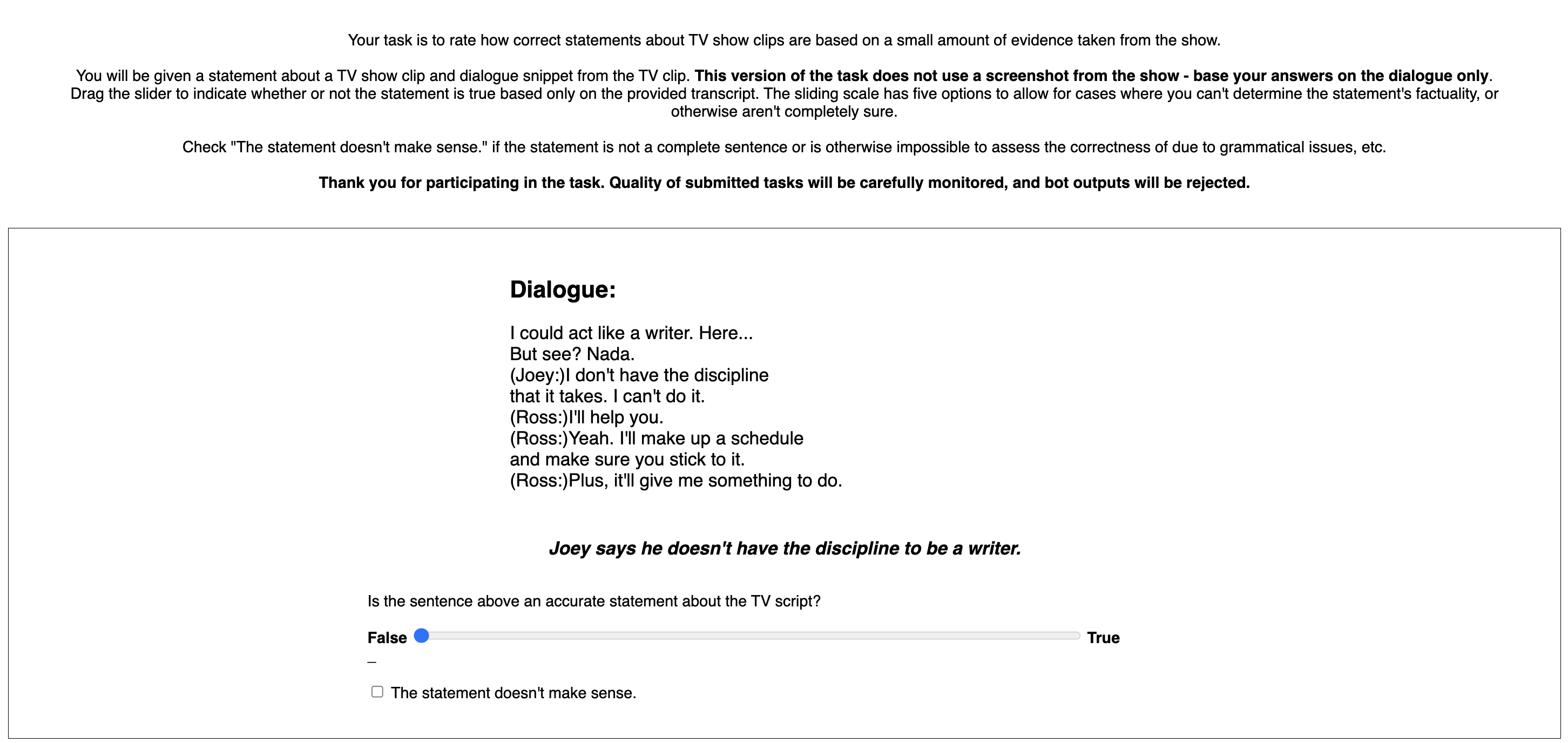}
  \caption{AMT acceptability task instructions and example for premises with textual evidence.}
\end{figure}

\begin{figure}
  \includegraphics[width=\textwidth]{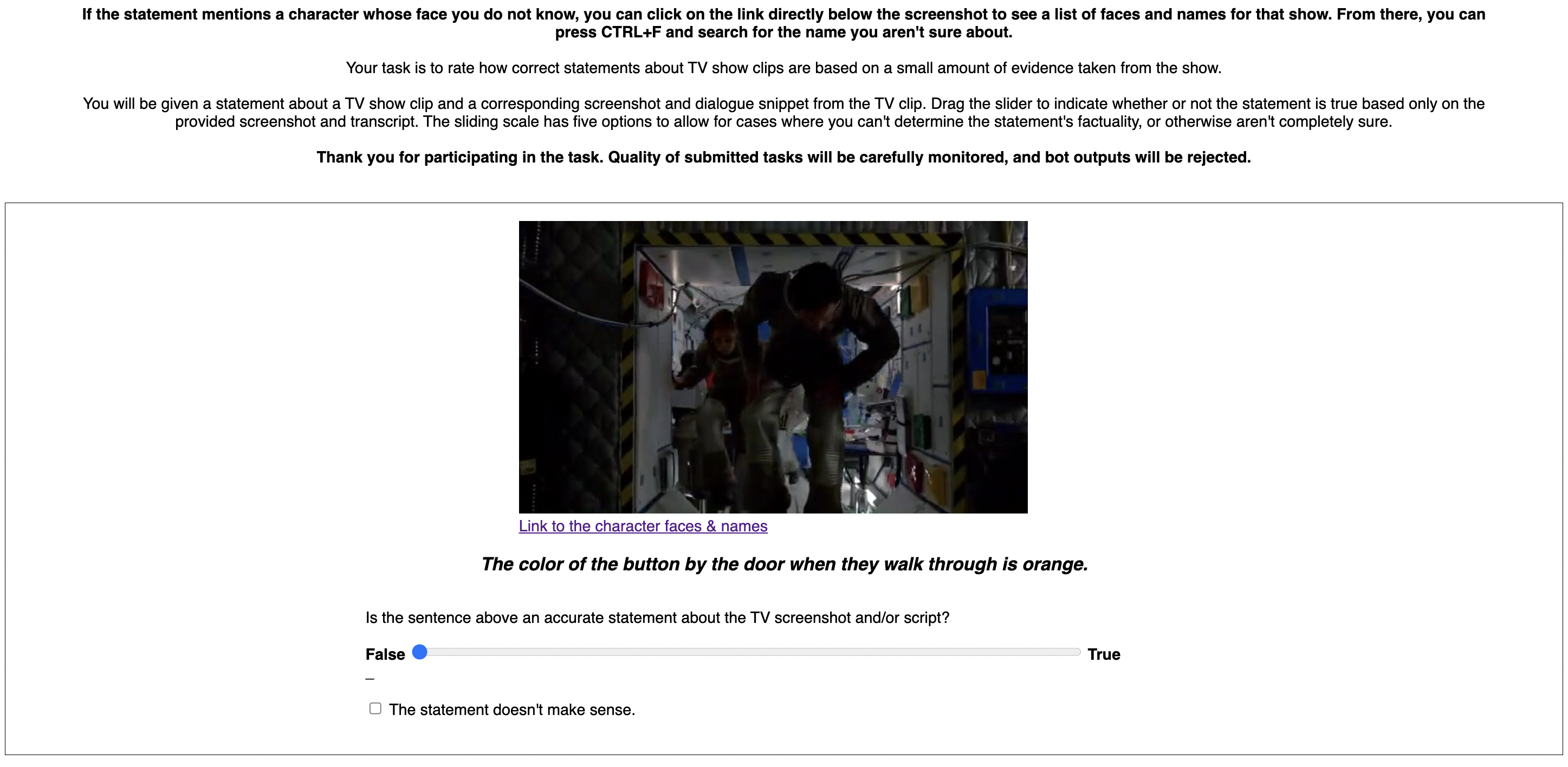}
  \caption{AMT acceptability task instructions and example for premises with visual evidence.}
\end{figure}

\noindent\textbf{Relevance:} See Figure 6.\\
\begin{figure}
  \includegraphics[width=\textwidth]{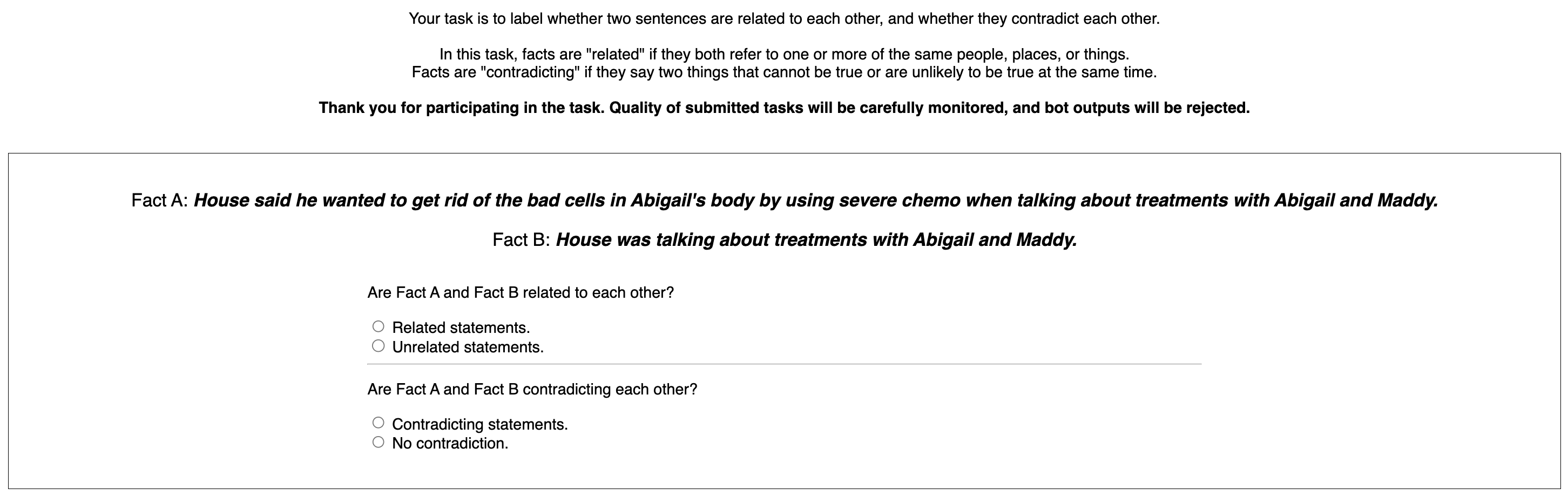}
  \caption{AMT relevance task instructions and example.}
\end{figure}

\noindent\textbf{Sufficiency:} See Figures 7 and 8.\\
\begin{figure}
  \includegraphics[width=\textwidth]{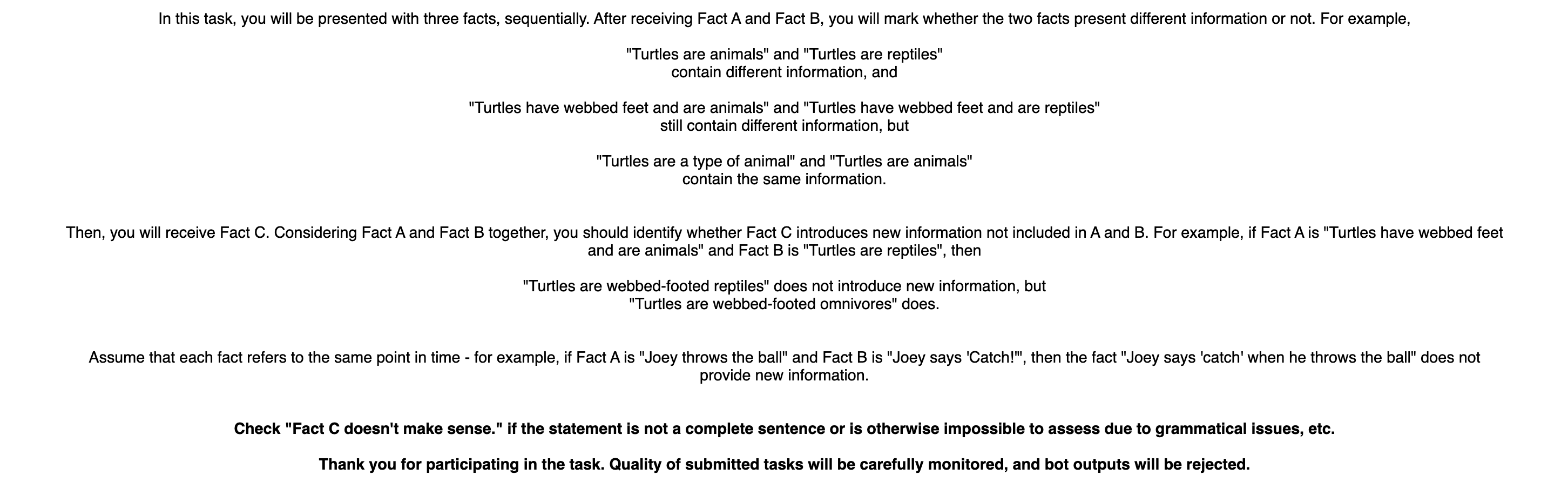}
  \caption{AMT sufficiency task instructions.}
\end{figure}

\begin{figure}
  \includegraphics[width=\textwidth]{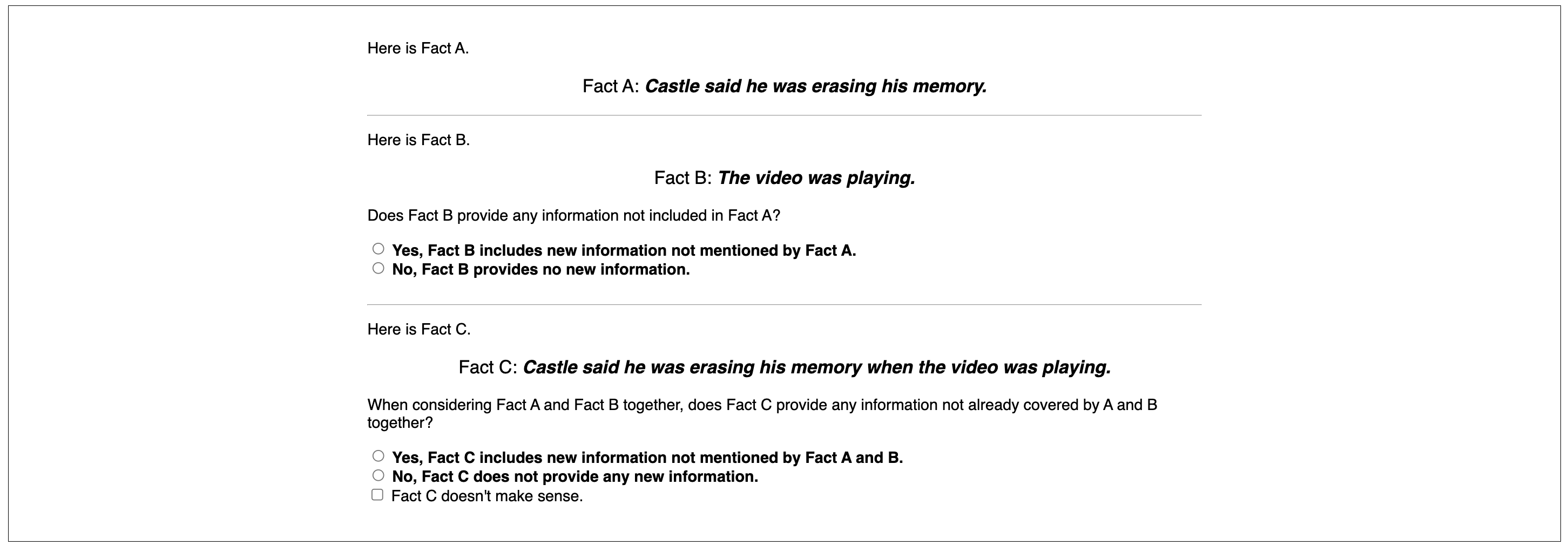}
  \caption{AMT sufficiency task example.}
\end{figure}

\newpage

\begin{table*}
    \centering
    \begin{tabular}{lcccccc}
    \Xhline{4\arrayrulewidth}
        \textbf{Method} & \textbf{FrozenBiLM} & \textbf{SeVILA} & \textbf{VideoChat2} & \textbf{TV-TREES$^\text{‡}$} & \textbf{TV-TREES} & \textbf{TV-TREES*}\\
        \toprule
        TVQA Acc.  & 26.3 & 38.2 & 40.6 & 44.9 & 49.4 & 48.1 \\
    \Xhline{4\arrayrulewidth}
    \end{tabular}
    \caption{Table contextualizing the anonymized VQA inputs ablation experiment (TV-TREES*) by comparing it to the other zero-shot TVQA results.}
    \label{tab:extra}
\end{table*}

\begin{table*}
    \centering
    \begin{tabular}{ll}
    \Xhline{4\arrayrulewidth}
        \textbf{Score} & \textbf{Description} \\
        \toprule
        \textbf{1} & Sentence is contradicted by the screenshot or dialogue.      \\
        \textbf{2} & Sentence is \textbf{unlikely} to be true based on the screenshot or dialogue.        \\
        \textbf{3} & Sentence is purely ambiguous given the screenshot or dialogue.      \\
        \textbf{4} & Sentence is \textbf{likely} to be true based on the screenshot or dialogue.       \\
        \textbf{5} & Sentence is directly suggested or shown by the screenshot or dialogue.      \\
    \Xhline{4\arrayrulewidth}
    \end{tabular}
    \caption{Descriptions for each acceptability score provided to annotators as part of the sliding bar functionality in the task.}
    \label{tab:amt}
\end{table*}



\input{prompts/hypothesis_gen}

\input{prompts/question_generation}

\input{prompts/branching}

\input{prompts/inference_generation}

\input{prompts/entailment_check}

\input{prompts/q_anonymization}

\input{prompts/entailment_h_check}

\input{prompts/llava_input}

\input{prompts/tree_eval_main}

\input{prompts/tree_eval_correctness}

\input{prompts/tree_eval_gpt4}



%% file: prompts/hypothesis_gen.tex
\begin{figure*}

\paragraph{Hypothesis Generation Prompt\\~\\} 

\footnotesize
{\tt
Convert each of the answer options for the following questions into GRAMMATICAL ANSWER SENTENCES. Make sure that they are FULL and COMPLETE sentences, not just words. They should be sentences that you can "prove" by reasoning about the situation.  Proving the sentence should amount to choosing choosing that answer option over the other one(s). \\~\\ \#\# Input \\QUESTION:\\ \{ICL Q Examples\}\\~\\ \#\# Output\\ \{ICL A Examples\} \\~\\~\\ \#\# Input\\QUESTION:\\ \{Questions\} \\~\\ \#\# Output\\}

\caption{Example prompt for generating hypotheses from QA pairs as described in Section 4.2.}

\end{figure*}

%% file: prompts/question_generation.tex
\begin{figure*}

\paragraph{Hypothesis-To-Question Generation Prompt\\~\\} 

\footnotesize
{\tt
Rewrite the following statement into a "yes" or "no" question, and nothing else.\\

STATEMENT: "\{Statement\}"\\
QUESTION: \\
}

\caption{Example prompt for generating interrogative forms of hypotheses for conditioning inference generation and VQA as described in Section 4.3.}

\end{figure*}

%% file: prompts/branching.tex
\begin{figure*}
\label{fig:fact_generator_prompt}

\paragraph{Hypothesis Decomposition Prompt\\~\\} 

\footnotesize
{\tt
You are a writing system that values clarity above all else. You NEVER uses pronouns like "he", "they", or "it" to ensure that readers can understand your sentences in isolation without additional context.\\

Your task is to break down the following statement into two, simpler sentences.\\

STATEMENT: "Lauren closed the door after discussing the party with Kelly."\\

DECOMPOSITION (USING NO PRONOUNS, INCLUDING "THEY" OR "HE" OR "SHE"):\\
(1) "Lauren closed the door."\\
(2) "Lauren discussed the party with Kelly."\\

STATEMENT: "Jason asked about the brown briefcase because he was concerned that it had been misplaced or stolen."\\

DECOMPOSITION (USING NO PRONOUNS, INCLUDING "THEY" OR "HE" OR "SHE"):\\
(1) "Jason asked about the brown briefcase."\\
(2) "Jason was concerned that the brown briefcase had been misplaced or stolen."\\

STATEMENT: "\{Statement\}"\\

DECOMPOSITION (USING NO PRONOUNS, INCLUDING "THEY" OR "HE" OR "SHE"):
}

\caption{Example prompt for decomposing a hypothesis into two distinct premises as described in Section 4.5.}

\end{figure*}

%% file: prompts/inference_generation.tex
\begin{figure*}

\paragraph{Inference Generation Prompt\\~\\} 

\footnotesize
{\tt
You are a fact-checking expert that uses evidence to answer questions about a TV show.\\~\\For the following question and scene dialogue, write a set of five independent inferences entailed by some part of the scene. The inferences should resemble short, factual statements about the scene and should help to answer the question using component reasoning steps.\\~\\Write your facts in JSON format, i.e. \{"1": "<answer here>", "2": "<answer here>", ...\} and nothing else.\\~\\QUESTION: "Why does Howard say they\'re late after walking in?" \\~\\SCENE: \\\{Dialogue\}\\~\\INFERENCES (5 total):\\
}

\caption{Example prompt for generating inferences from dialogue samples given an underlying question as described in Section 4.3.}

\end{figure*}

%% file: prompts/entailment_check.tex
\begin{figure*}

\paragraph{Premise-Dialogue Entailment Verification Filtering Prompt\\~\\} 

\footnotesize
{\tt
You are an expert social reasoning system that understands the implied meanings of complex conversations between TV show characters. Given social inferences made by other AI systems about transcripts, you score them on whether they are CORRECT or NOT SUPPORTED by the transcript.\\~\\Given the following TV show transcript, write whether each of the following statements about the TV show are CORRECT or NOT SUPPORTED. A statement is CORRECT if an average human would agree that it is most likely true based on the transcript, and is NOT SUPPORTED otherwise.\\~\\Write your facts in JSON format, i.e. \{"1": <"answer here">, "2": <"answer here">, ...\} and nothing else.\\~\\TRANSCRIPT:\\ \{Dialogue\} \\~\\STATEMENTS:\\\{Inferences\}\\~\\OUTPUT:\\
}

\caption{Example prompt for filtering premises based on dialogue entailment as described in Section 4.3.}

\end{figure*}

%% file: prompts/q_anonymization.tex
\begin{figure*}

\paragraph{Question Anonymization Prompt\\~\\} 

\footnotesize
{\tt
Anonymize the following questions by replacing all the characters' names replaced with \"the man\", \"the woman\", \"the person\", or \"the people\". Your output should be formatted as a serialized JSON list, i.e. \{\"q1\": \"<answer here>\", \"q2\": \"<answer here>\"\}, ..., and nothing else.\\~\\SENTENCES:\\\{Questions\}\\~\\QUESTIONS:\\
}

\caption{Example prompt for generating anonymized versions of interrogative versions of hypotheses as described in Appendix B.}

\end{figure*}

%% file: prompts/entailment_h_check.tex
\begin{figure*}

\paragraph{Premise-Hypothesis Entailment Verification Filtering Prompt\\~\\} 

\footnotesize
{\tt
You are a logical reasoning system that determines whether individual facts are enough to prove a hypothesis statement.\\~\\For each of the following independent facts, answer "YES" if the fact cannot be true without the hypothesis also being true, and "NO" if the hypothesis can be false even if the fact is true. Always answer "NO" if the hypothesis is not a complete sentence (for example "is sitting.". Write your answers in JSON format, i.e. \{"1": "<fact 1 answer here>", "2": "<fact 2 answer here>", ...\} and nothing else.\\~\\HYPOTHESIS: \{Hypothesis\}\\~\\FACTS:\\\{Inferences\}\\~\\OUTPUT: \\
}

\caption{Example prompt for filtering premises based on hypothesis entailment  as described in Section 4.4.}

\end{figure*}

%% file: prompts/llava_input.tex
\begin{figure*}

\paragraph{Visual QA Prompt\\~\\} 

\footnotesize
{\tt
From this image, can you answer the question \{Question\}? If so, answer the question, otherwise, answer \"NOT ENOUGH INFO\".
}

\caption{Prompt template for soliciting VQA outputs from the LLaVA-7B model as described in Section 4.6.}

\end{figure*}

%% file: prompts/tree_eval_main.tex
\begin{figure*}

\paragraph{GPT-4 Relevance, Distinctness, and Sufficiency Evaluation\\~\\} 

\footnotesize
{\tt
You are a reasoning system that searches for proofs of a hypothesis about a video clip by recursively decomposing it into simpler premises.\\

Given a hypothesis, you identify entries in a list of possible two-premise decompositions of the hypothesis that are “well-formed”: Proving the premises of a well-formed decomposition would amount to proving the hypothesis through compositional entailment.\\

You assess decompositions using three metrics: Premise relevancy, premise distinctness, and decomposition sufficiency. Each decomposition should receive two relevancy and distinctness scores, one for each premise, but only one single sufficiency score.\\

RELEVANCY: Relevancy measures whether a premise contributes information pertaining to the hypothesis. This is measured on a binary scale. Simply, if the premise mentions an entity or idea also mentioned by the hypothesis, the relevancy score is 1. Otherwise, it is 0.\\

DISTINCTNESS: Distinctness measures whether a premise introduces new information not already entailed by the other premise in the decomposition. This is measured on a binary scale. If the premise only introduces information already entailed by the other premise in the decomposition, the distinctness score is 0. Otherwise, it is 1. If both premises are the same, both receive a score of 0.\\

SUFFICIENCY: Sufficiency measures whether the two premises cover all the information introduced by the hypothesis. This is also measured on a binary scale. If, when considering both premises, the hypothesis introduces new information not covered by the decompositional premises, the sufficiency score is 0. If the hypothesis does not introduce new information, the sufficiency score is 1.\\

For the following decompositions, score each decomposition’s relevancy and sufficiency. Decompositions will be presented in the form “(<decomposition number>) H: <hypothesis> \& P1: <decomp premise 1> \& P2: <decomp premise 2>”. Your answer should be a list of entries taking the form “(<decomposition number>) RELEVANCY: (<premise 1 score>, <premise 2 score>), DISTINCTNESS: ((<premise 1 score>, <premise 2 score>), SUFFICIENCY: (<overall score>)”.\\

DECOMPOSITIONS:\\
\{Decompositions\}\\

JUDGEMENTS (one line per decomposition):

}

\caption{GPT-4 prompt for scoring the relevance, distinctness, and sufficiency of decompositions in an entailment tree.}

\end{figure*}

%% file: prompts/tree_eval_correctness.tex
\begin{figure*}

\paragraph{GPT-4 Textual Acceptability Evaluation\\~\\} 

\footnotesize
{\tt
Based on the dialogue from the TV show, how likely is it that the statements below are true? Score the likelihood of each statement on a 1-5 scale, where 1 indicates the dialogue contradicts the statement, 2 indicates the statement is unlikely to be true given the dialogue, 3 indicates the statement is ambiguous given the dialogue, 4 indicates the statement is likely to be true given the dialogue, and 5 indicates that the statement must be true given the dialogue. Write your numerical scores in the same order as the listed statements, separated by commas, and nothing else. \\ 

Dialogue: \\
\{Dialogue\}  \\~\\
Statements:\\
\{Statements\}
}

\caption{GPT-4 prompt for scoring the acceptability of entailment tree leaf nodes that cite textual evidence.}

\end{figure*}

%% file: prompts/tree_eval_gpt4.tex
\begin{figure*}

\paragraph{GPT-4V Visual Acceptability Evaluation\\~\\} 

\footnotesize
{\tt
Based on the screenshot from the TV show, how likely is it that the statement below is true? Score the likelihood on a 1-5 scale, where 1 indicates the screenshot contradicts the statement, 2 indicates the statement is unlikely to be true given the screenshot, 3 indicates the statement is ambiguous given the screenshot, 4 indicates the statement is likely to be true given the screenshot, and 5 indicates that the statement must be true given the screenshot. Write your numerical score and nothing else. \\

Statement: \{Statement\}
}

\caption{GPT-4V prompt for scoring the acceptability of entailment tree leaf nodes that cite visual evidence. The top-scoring video frame is passed in alongside the prompt.}

\end{figure*}